# An Exploratory Study of Tweets about the SARS-CoV-2 Omicron Variant: Insights from Sentiment Analysis, Language Interpretation, Source Tracking, Type Classification, and Embedded URL Detection

**Nirmalya Thakur [1]* and Chia Y. Han [2]**

[1] Department of Electrical Engineering and Computer Science, University of Cincinnati, Cincinnati, OH 45221-0030, U.S.A.; thakurna@mail.uc.edu
[2] Department of Electrical Engineering and Computer Science, University of Cincinnati, Cincinnati, OH 45221-0030, U.S.A.; han@ucmail.uc.edu
* Correspondence: thakurna@mail.uc.edu

**Abstract:** This paper presents the findings of an exploratory study on the continuously generating Big Data on Twitter related to the sharing of information, news, views, opinions, ideas, knowledge, feedback, and experiences about the COVID-19 pandemic, with a specific focus on the Omicron variant, which is the globally dominant variant of SARS-CoV-2 at this time. A total of 12028 tweets about the Omicron variant were studied, and the specific characteristics of tweets that were analyzed include - sentiment, language, source, type, and embedded URLs. The findings of this study are manifold. First, from sentiment analysis, it was observed that 50.5% of tweets had a 'neutral' emotion. The other emotions - 'bad', 'good', 'terrible', and 'great' were found in 15.6%, 14.0%, 12.5%, and 7.5% of the tweets, respectively. Second, the findings of language interpretation showed that 65.9% of the tweets were posted in English. It was followed by Spanish or Castillian, French, Italian, Japanese, and other languages, which were found in 10.5%, 5.1%, 3.3%, 2.5%, and <2% of the tweets, respectively. Third, the findings from source tracking showed that "Twitter for Android" was associated with 35.2% of tweets. It was followed by "Twitter Web App", "Twitter for iPhone", "Twitter for iPad", "TweetDeck", and all other sources that accounted for 29.2%, 25.8%, 3.8%, 1.6%, and <1% of the tweets, respectively. Fourth, studying the type of tweets revealed that retweets accounted for 60.8% of the tweets, it was followed by original tweets and replies that accounted for 19.8% and 19.4% of the tweets, respectively. Fifth, in terms of embedded URL analysis, the most common domain embedded in the tweets was found to be twitter.com, which was followed by biorxiv.org, nature.com, wapo.st, nzherald.co.nz, recvprofits.com, science.org, and other domains. Finally, to support similar research and development in this field centered around the analysis of tweets, we have developed an open-access Twitter dataset that comprises more than 500,000 tweets about the SARS-CoV-2 omicron variant since the first detected case of this variant on November 24, 2021.

**Keywords:** COVID-19; SARS-CoV-2; Omicron; Twitter; Tweets; Sentiment Analysis; Big Data; Natural Language Processing; Data Science; Data Analysis

## 1. Introduction

An outbreak of an unknown respiratory disease started in December 2019 in the seafood market in Wuhan, China, infecting about 66% of the people at the market. Very soon, more people in different parts of China got infected by the same disease. This prompted an investigation from the healthcare and medical sectors, and very soon, it was concluded that a novel coronavirus was responsible for this disease. This novel coronavirus was named severe acute respiratory syndrome coronavirus-2 (SARS-CoV-2, 2019-nCoV) as it was observed to have a high homology of about 80% with SARS-CoV [1]. The disease that humans suffer from after getting infected by this virus is known as COVID-19 [2]. The Chinese government implemented multiple measures to reduce the spread of the virus. However, the virus rapidly spread across China and soon started spreading in different countries of the world. COVID-19 was declared a pandemic by the World Health Organization (WHO) on March 11, 2021 [4].

At the time of writing this paper, globally, there have been 549,667,293 cases of COVID-19 with 6,352,025 deaths [5]. The SARS-CoV-2 virus mainly attacks the respiratory system of humans, although infections in other organs of the body have also been reported in some cases. The symptoms, as reported from the initial studies of the cases from Wuhan, China, include fever, dry cough, dyspnea, headache, dizziness, exhaustion, vomiting, and diarrhea. The report also mentions that not everyone has the same symptoms, and the nature and intensity of the symptoms vary from person to person [6]. When the genetic sequence of a virus changes, it is said to have mutated. Genomes of a virus that are different from each other in terms of their genetic sequence are called variants. Variants that differ in terms of phenotype are known as strains [7]. On January 10, 2020, the genome sequences of SARS-CoV-2 were made publicly available by the Global Influenza Surveillance and Response System (GISAID), which is a primary source on a global scale for open access to the genomic data of influenza viruses [8]. Since then, the database of GISAD has made more than 5 million genetic sequences of SARS-CoV-2 from 194 countries and territories publicly available for research [9]. In an attempt to prioritize research related to COVID-19, the WHO has made a conscious effort to classify the variants of SARS-CoV-2 into three categories. These include variants of concern (VOCs), variants of interest (VOIs), and variants under monitoring (VUMs).

Since the initial cases in December 2019, the SARS-CoV-2 virus has undergone multiple mutations, and as a result, several variants have been detected in different parts of the world. Some of these include – Alpha (B.1.1.7), Beta (B.1.351), Gamma (P.1), Delta (B.1.617.2), Epsilon (B.1.427 B.1.429), Eta (B.1.525), Iota (B.1.526), Kappa (B.1.617.1), Zeta (P.2), Mu (B.1.621, B.1.621.1), and Omicron (B.1.1.529, BA.1, BA.1.1, BA.2, BA.3, BA.4 and BA.5) [10]. Out of all these variants, the Omicron variant, first detected on November 24, 2021, was classified as a VOC. by WHO on November 26, 2021 [11]. Soon thereafter, the Omicron variant became the globally dominant form of SARS-CoV-2 [12]. The Omicron spike protein contains 30 mutations and has been reported to be the most immune evasive VOC of SARS-CoV-2 and is highly resistant against antibody-mediated neutralization [13,14]. Despite the development of vaccines [15] and various forms of treatments [16], a recent report from WHO states that the Omicron variant accounts for 86% of the infections caused by SARS-CoV-2 on a global scale [17]. In another recent report, WHO mentioned that the cases due to the Omicron variant were "off the charts" as global infections due to the SARS-CoV-2 variant set new records [18]. Currently, some of the countries that have recorded the most cases due to the SARS-CoV-2 Omicron variant include – United Kingdom (1,138,814 cases), USA (945,470 cases), Germany (245,120 cases), Denmark (218,106 cases), France (110,959 cases), Canada (92,341 cases), Japan (71,056 cases), India (56,125 cases), Australia (46,576 cases), Sweden (43,400 cases), Israel (39,908 cases), Poland (33,436 cases), and Brazil (32,880 cases) [19].

In today's world, where the internet virtually connects people in different geographic regions, social media has been considered an "integral vehicle" of people's lives and an "online community" for communication, information, news, views, opinions, perspectives, ideas, knowledge, feedback, and experiences related to pandemics, epidemics, viruses, and diseases [20-24]. Out of all the social media platforms, Twitter is popular amongst all age groups [25], and there are about 192 million daily active users on Twitter [26].

Twitter has been highly popular amongst healthcare researchers, epidemiologists, medical practitioners, data scientists, and computer science researchers for studying, analyzing, modeling, and interpreting social media communications related to pandemics, epidemics, viruses, and diseases such as Ebola [27], E-Coli [28], Dengue [29], Human papillomavirus (HPV) [30], Middle East Respiratory Syndrome (MERS) [31], Measles [32], Zika virus [33], H1N1 [34], influenza-like illness [35], swine flu [36], flu [37], Cholera [38], Listeriosis [39], cancer [40], Liver Disease [41], Inflammatory Bowel Disease [42], kidney disease [43], lupus [44], Parkinson's [45], Diphtheria [46], and West Nile virus [47].

Comprehensive analysis of multimodal aspects of tweets posted on Twitter has been of significant interest to the scientific community during similar epidemics and virus outbreaks in the past. For instance, during the outbreak of the Ebola Virus, researchers studied tweets posted during the outbreak to perform sentiment analysis [48], embedded URL detection [49], tweet content investigation [50], and analysis of the tweet type, such as retweets [51]. Similarly, during the outbreak of the Zika virus, researchers studied tweets posted during the outbreak to perform sentiment analysis [52], investigate retweet characteristics [53], detect embedded URLs in tweets [54], understand the source of the tweets [55], and detect the language of the tweets [56]. A third example can be seen from the research works related to studying the multimodal aspects of tweets that focused on the flu outbreak. Relevant tweets posted both during and just prior to the outbreak were studied for performing sentiment analysis [57], tweet content analysis [58], studying retweet characteristics [59], and embedded word analysis [60].

The outbreak of the COVID-19 pandemic has served as a "catalyst" towards increasing the usage of Twitter [61-64] for conversations on a wide range of topics in this regard resulting in tremendous amounts of Big Data. Some of the most popular use cases of Twitter during this pandemic, as reported in recent works in this field, include the following:

1. Sharing of symptoms, information, and experiences as reported by frontline workers and people who were infected with the virus [65]
2. Providing suggestions, opinions, and recommendations to reduce the spread of the virus [66]
3. Communicating updates on vaccine development, clinical trials, and other forms of treatment [67]
4. Sharing guidelines mandated by various policy-making bodies, such as mask mandate, social distancing, etc. that were required to be followed by members in specific geographic regions of the world under the authority of the associated policy-making bodies [68]
5. Dissemination of misinformation such as the use of certain drugs or forms of treatment that have not been tested or have not undergone clinical trials [69]
6. Creating and spreading conspiracy theories such as considering 5G technologies responsible for the spread of COVID-19, which eventually led to multiple 5G towers being burnt down in the United Kingdom [70]
7. Studying public opposition to available vaccines in different parts of the world [71].

In addition to these works related to the use of Twitter during COVID-19, there have been some other works in this field (as discussed in detail in Section 2) that further uphold the extensive increase in Twitter posts for information seeking and sharing during the ongoing surge of COVID-19 cases due to the Omicron variant. However, none of these works in this field took a comprehensive approach towards drawing insights from such tweets - such as performing sentiment analysis, tweet type detection, embedded URL tracking, tweet source tracking, and language detection that were performed during the outbreaks of viruses in the past such as - during the outbreaks of Ebola Virus [48-51], Zika virus [52-56], and the flu outbreak [57-60]. To address this research challenge, in this work, a total of 12,028 recent tweets containing views, expressions, opinions, perspectives, attitudes, news, information, and related themes about the SARS-CoV-2 Omicron variant posted publicly on Twitter between May 5, 2022, to May 12, 2022, were studied, analyzed, and interpreted to perform a comprehensive analysis of the associated tweets. In summary, the findings of this study are as follows:

1. The results from sentiment analysis showed that a majority of the tweets (50.5%) had a 'neutral' emotion, which was followed by the emotional states of 'bad', 'good', 'terrible', and 'great' found in 15.6%, 14.0%, 12.5%, and 7.5% of the tweets, respectively.
2. The results from Tweet source tracking showed that 35.2% of the Tweets were posted from an android source, which was followed by the Twitter Web App, iPhone, iPad, Tweetdeck, and other sources, which accounted for 29.2%, 25.8%, 3.8%, 1.6%, and <1% of the tweets, respectively
3. The results from Tweet language interpretation showed that 65.9% of the tweets were posted in English, which was followed by Spanish or Castillian (10.5%), French (5.1%), Italian (3.3%), Japanese (2.5%), and other languages that accounted for <2% of the tweets.
4. The results from Tweet type classification showed that the majority of the tweets (60.8%) were retweets which was followed by original tweets (19.8%) and replies (19.4%).
5. The results from embedded URL analysis showed that the most common domain embedded in the tweets was found to be twitter.com, which was followed by biorxiv.org, nature.com, wapo.st, nzherald.co.nz, recvprofits.com, science.org, and a few other sources.

In addition to the above, to further support research and development in this field, we present an open-access dataset of 537,702 Tweet IDs of the same number of tweets about the COVID-19 omicron variant, posted publicly on Twitter since the first detected case of this variant on November 24, 2021. The development of Twitter datasets has been of significant interest to the scientific community, as can be seen from the recent Twitter datasets on the 2020 U.S. Presidential Elections [72], 2022 Russia Ukraine war [73], climate change [74], natural hazards [75], European Migration Crisis [76], movies [77], toxic behavior amongst adolescents [78], music [79], civil unrest [80], drug safety [81], and Inflammatory Bowel Disease [82]. Since the outbreak of COVID-19, there have been a few works that focused on the development of Twitter datasets. These include an Arabic Twitter dataset [83], a Twitter dataset for vaccine misinformation [84], a Twitter dataset on misleading information about COVID-19 [85], a Twitter dataset for COVID-19-related misinformation [86], and a dataset on COVID-19 rumors [87]. However, none of these works [83-87] have focused on the development of a dataset that comprises tweets about the Omicron variant that were posted on Twitter since the first detected case of this variant. Therefore, the development of this dataset further helps to uphold the scientific contributions of this paper. This open-access dataset, publicly available at https://doi.org/10.5281/zenodo.6804323, is compliant with the privacy policy, developer agreement, and guidelines for content redistribution of Twitter, as well as with the FAIR principles (Findability, Accessibility, Interoperability, and Reusability) principles for scientific data management. This paper is presented as follows. In Section 2, a review of recent works in this field is presented. The methodology is discussed in Section 3. Section 4 discusses the results. The conclusion is presented in Section 5, which summarizes the scientific contributions of this study and outlines the scope for future work in this field. It is followed by the references section.

## 2. Literature Review

In this section, a review is presented of the recent works in this field that focused on studying social media behavior with a specific focus on Twitter since the outbreak of SARS-CoV-2. Shamrat et al. [88] developed a kNN-based machine learning classifier to classify tweets related to COVID-19 into three classes – 'positive', 'negative', and 'neutral'. The study specifically focused on filtering tweets related to COVID-19 vaccines, and thereafter this algorithm was applied to the filtered tweets for analysis. Sontayasara et al. [89] used the support vector machine (SVM) classifier to develop an algorithm for sentiment analysis. The algorithm was tested on tweets where people communicated their plans to visit Bangkok during the pandemic and how those plans were affected. This classifier was able to classify the tweets into three classes of sentiments - 'positive', 'negative', and 'neutral'. Asgari-Chenaghlu et al. [90] developed an approach to detect the trending topics and major concerns related to the COVID-19 pandemic as expressed by people on Twitter. Amen et al. [91] proposed a framework that applied a directed acyclic graph model on Tweets related to COVID-19 to detect any anomaly events. The work of Liu et al. [92] involved developing an approach to study tweets about COVID-19 that involved the Centers for Disease Control and Prevention (C.D.C.). The objective of this study was to detect public perceptions such as concerns, attention, expectations, etc., related to the guidelines of the C.D.C. regarding COVID-19. Al-Ramahi et al. [93] developed a methodology to filter and study tweets posted between January 1, 2020, and October 27, 2020, where people expressed their opposing views towards wearing masks to reduce the spread of COVID-19. Jain et al. [94] proposed a methodology to analyze tweets related to COVID-19 that could assign an influence score to the associated users who posted these tweets. The objective of this study was to identify influential users on Twitter who posted about COVID-19. Madani et al. [95] developed a random forest-based classifier to detect tweets about COVID-19 that contained fake news. The classifier achieved a performance accuracy of 79%.

Shokoohyar et al. [96] proposed a system to study tweets where people expressed their opinions regarding the lockdown in the United States on account of COVID-19. Chehal et al. [97] developed a software using Python and R to analyze the mindset of Indians as expressed in their tweets during the two nationwide lockdowns that were implemented by the Indian government on account of COVID-19. Glowacki et al. [98] developed a systemic approach to identify and study tweets related to COVID-19 where Twitter users discussed addiction issues. Selman et al. [99]'s study focused on studying tweets where Twitter users reported their relative, friend, or acquaintance passing away from COVID-19. The study specifically focused on patients who were reported to have been alone at the time of their death. The work of Koh et al. [100] aimed to identify tweets using specific keywords where Twitter users communicated about feelings of loneliness during COVID-19. The authors tested their approach on a total of 4492 tweets. Mackey et al. [101]'s work focused on filtering and investigating tweets related to COVID-19 where people self-reported their symptoms, access to testing sites, and recovery status. In [102], the authors focused on studying tweets related to COVID-19 to understand the anxiety and panic buying behavior with a specific focus on buying toilet paper during this pandemic. The work involved specific inclusion criteria for the tweets, and a total of 4081 tweets were studied.

As can be seen from these works involving studying social media behavior and user characteristics on Twitter during COVID-19, while there have been several innovations and advancements made in this field, the following limitations exist in these works:

1. Most of these works used approaches to detect tweets that contained one or more keywords, hashtags, or phrases such as "COVID-19", "coronavirus", "SARS-CoV-2", "covid", "corona," etc. but none of these works focused on including one or more keywords directly related to the SARS-CoV-2 omicron variant to include the associated tweets. As the SARS-CoV-2 Omicron variant is now responsible for most of the COVID-19 cases globally, the need in this context is to filter tweets that contain one or more keywords, hashtags, or phrases related to this variant.
2. The works on sentiment analysis [88, 89] focused on the proposal of new approaches to detect the sentiment associated with tweets; however, the categories for classification of the associated sentiment were only 'positive', 'negative', and 'neutral'. In a realistic scenario, there can be different kinds of 'positive' emotions, such as 'good' and 'great'. Similarly, there can be different kinds of 'negative' emotions, such as 'bad' and 'terrible'. The existing works cannot differentiate between these kinds of positive (or negative) emotions. Therefore, the need in this context is to expand the levels of sentiment classification to include the different kinds of positive and negative emotions.

3. While there have been multiple innovations in this field of Twitter data analysis - such as detecting trending topics [90], anomaly events [91], public perceptions towards C.D.C. [92], and views towards not wearing masks [93], just to name a few, there has been minimal work related to quantifying and ranking the associated insights.
4. The number of tweets that were included in previous studies (such as 4081 tweets in [102] and 4492 tweets in [100]) comprises a very small percentage of the total number of tweets that have been posted related to COVID-19 since the beginning of the outbreak. Therefore, the need in this context is to include more tweets in the studies.

The work proposed in this paper aims to explore the intersections of Big Data mining, Natural Language Processing, Data Science, Information Retrieval, Machine Learning, and their related areas to address the above-mentioned needs. The methodology is outlined in the next section.

## 3. Materials and Methods

This section is divided into multiple parts. Section 3.1 provides an overview of how this work is compliant with the privacy policy [103] and developer agreement [104] of Twitter. Section 3.2 provides a brief overview of the research tool that was used for this study. Section 3.3 presents the methodology of the work that was performed. Section 3.4 outlines the approach that was followed for the development of the associated Twitter dataset.

### *3.1. Compliance with Twitter policies*

The work primarily involves drawing insights by mining tweets. Therefore, the privacy policy [103] and developer agreement and policy [104] of Twitter were studied at first. The privacy policy of Twitter [103] states – *"Twitter is public and Tweets are immediately viewable and searchable by anyone around the world"*. It also states – *"Most activity on Twitter is public, including your profile information, your display language, when you created your account, and your Tweets and certain information about your Tweets like the date, time, and application and version of Twitter you Tweeted from…..The lists you create, people you follow and who follow you, and Tweets you Like or Retweet are also public……..By publicly posting content, you are directing us to disclose that information as broadly as possible, including through our APIs, and directing those accessing the information through our APIs to do the same."* To add, the Twitter developer agreement and policy [104] defines tweets as *"public data"*. Therefore, based on the terms and conditions mentioned in these two policies, it can be concluded that performing this data analysis and drawing insights from tweets by connecting with the Twitter API is compliant and adheres to both these policies.

### *3.2. Overview of Social Bearing*

The work was performed by using Social Bearing, a research tool for performing Twitter research [105]. It was developed by Tom Elliott, and the tool was made available for the public starting on January 1, 2015. The tool uses multiple JavaScript, text processing, and text analysis libraries and algorithms in combination as well as in standalone form for performing Big Data Mining, Data Analysis, Information Processing, and Information Retrieval on tweets (obtained based on keyword or hashtag search from Twitter). The specific JavaScript libraries used by this research tool include - jquery-ui.min.js [106], jquery.tinysort.min.js [107], masonry.pkgd.min.js [108], d3.layout.cloud.js [109], d3.min.js [110], analytics.js [111], and loader.js [112].

This is a popular tool amongst academic researchers in this field and has been used for several interdisciplinary applications that focused on studying and analysis of Tweets in the last few years [113-121]. These include – analysis of the literacy of the Twitter metaverse [113], preparing Twitter data for a MANCOVA Test [114], studying the usage of Twitter to reduce drunk driving on New Year's Eve [115], tracking fake news related to COVID-19 vaccinations [116], studying polarized politics communicated via Twitter in India [117], assessing social support and stress in autism-focused virtual communities on Twitter [118], tracking online propaganda by Twitter usage [119], investigating the proliferation of Twitter accounts in a Higher Education setting [120], and studying market identity claims on Twitter [121].

There are several advantages of this research tool as compared to several other research tools which were used in prior works in this field. These are outlined as follows:

1. The tool works by complying with the privacy policy, developer agreement, and developer policies of Twitter [103,104] and does not constitute any illegal, unethical, or unauthorized calling, posting, or scraping performed on the Twitter API [122].
2. The traditional approach of searching Tweets based on keyword search involves a series of steps which include - setting up a Twitter developer account, obtaining the GET oauth/authorizes code, obtaining the GET oauth/authenticate code, obtaining the Bearer Token, and obtaining the Twitter API Access Token and Secret and manually entering the same in the given application (such as a Python program) to connect with the Twitter API. As this research tool is already set up to work in accordance with the privacy policy, developer agreement, and developer policies of Twitter, therefore, the manual entry of all the above-mentioned codes and tokens is not required, and the process of connecting with the Twitter API to search tweets is simplified. It can be performed by just clicking the Twitter Sign In button on the visual interface, and then after the user signs into Twitter with an active Twitter account, the tool is set up to work by searching tweets based on the keyword(s) or the hashtag(s).
3. The tool uses the in-built JavaScript, text processing, and text analysis libraries and algorithms to extract characteristics from the relevant tweets and displays the same on the visual interface in the form of results, thereby reducing the time and complexity of developing these algorithms and functions from scratch as well as for developing the approaches for visualizing the results.
4. While displaying the results, the tool shows a list of users (with their Twitter usernames) whose tweets were fetched in the "All Contributors" section of the tool. It allows removing one or more users from this list (if the usernames indicate that the Twitter profiles are bots) so that the new results are obtained based on the tweets posted by the rest of the users.

In addition to the above characteristic features that uphold the applicability of this research tool for different use cases, another reason for the popularity of this research tool can be attributed to the fact that no software development or API development or library development is necessary when is tool is used for a study such as this one. The users of this tool are only required to be familiar with the functionalities of the tool and how to use the same for collecting, studying, and analyzing Twitter data. As this research tool was used for this study, so this paper does not report any code, program, software, library, or API as a supplementary item on any open-source repository such as GitHub.

### *3.3. Methodology*

As can be seen from Figure 1, the first step was to perform Big Data mining to collect these tweets from Twitter. The Twitter search API has a 7-day limit on the tweets that can be searched [123]; in other words, tweets posted more than 7-days ago cannot be searched and studied. So, a total of 12028 recent tweets about the SARS-CoV-2 Omicron variant were studied that were posted on Twitter between May 12, 2022 (the most recent date at the time of submission of this paper) and May 5, 2022 (the date up to which tweets could be searched in compliance with the Twitter standard search API guidelines). This data mining process was performed by filtering tweets from Twitter in this date range that contained the “omicron” keyword or hashtag. This was performed to capture tweets that comprised views, expressions, opinions, perspectives, attitudes, news, information, and similar conversation themes related to the SARS-CoV-2 omicron variant, where other than the “omicron” keyword, the rest of the words used in the respective tweets would be associated with such information.

Multiple random sub-samples of the data (consisting of 7 tweets in each sample) were studied to evaluate the commonly used phrases to refer to the SARS-CoV-2 omicron variant. These phrases included – “first case of omicron variant”, “approaching its Omicron peak”, “Omicron is a stone cold killer”, “case of Omicron sub-variant”, “new omicron sub-variants”, “the highly transmissible omicron variant”, “sinus pain of omicron”. Even though these phrases are different, the keyword “omicron” is present in all these phrases. This helps to uphold the relevance of performing the data mining and data collection of relevant tweets by using this keyword. The process of data mining that was followed was not case sensitive, so the tweets containing the “omicron” keyword (or hashtag) as “omicron” or “Omicron” or “OMICRON” or any other order of capitalization of these alphabets were also filtered, and these specific cases did not need to be specifically mentioned as separate or different keywords (or hashtags).

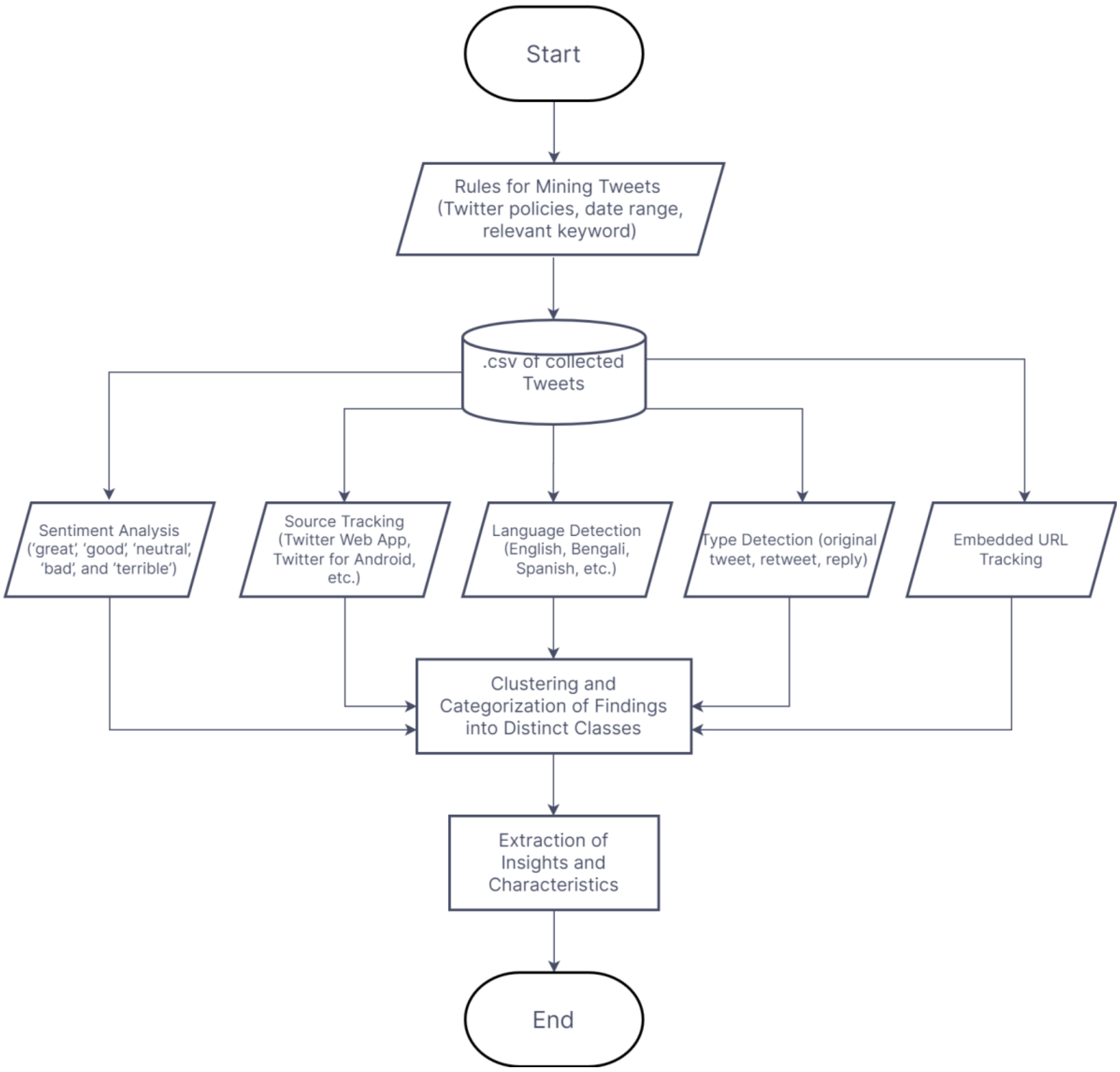

**Figure 1.** Flowchart-based representation of the methodology that was followed.

For all these tweets (available as a .csv file after the data mining process was completed), the specific insights that were studied, computed, interpreted, analyzed, quantified, and ranked in the next steps included the sentiment, source tracking, detecting the language of the tweet, inferring the type of tweet, and detecting the URLs embedded in the tweets. The Social Bearing tool's visual interface shows all the Twitter user profiles whose tweets are being considered for a specific use-case scenario for which the tool is being used. The Twitter usernames of these profiles are listed in the "All contributors" section. Prior to performing this comprehensive analysis, the "All contributors" section of the tool was thoroughly reviewed and the accounts that had Twitter handles or usernames which indicated that those accounts are likely to be bots were manually unselected. As a result, the tweets from these bot accounts were not included for the analysis. In addition to this, in the "Other Options" section on Social Bearing's visual interface, there is a checkbox – "Remove Duplicate Tweets". We had checked this checkbox during the study and the tool automatically removed the duplicate tweets from the results. Or in other words duplicate tweets did not play a role in the findings that are presented. The process of sentiment analysis involved tokenization and lemmatization of the text of the tweet and classifying it into various sentiments such as 'great', 'good', 'neutral', 'bad', and 'terrible'. While there can be multiple ways of performing sentiment analysis, this process of sentiment analysis was selected as multiple sentiment classes can be created, and the resultant classification is neither a binary classification between 'positive' and 'negative' sentiments nor is it a 3 level classification between 'positive', 'negative' and 'neutral' sentiment classes, which have been the limitations in prior works in this field (discussed in Section 2). The source tracking aspect of the study involved tracking the publicly available "source label" [124] that Twitter associates with each tweet. This source level has different categories, which include - Twitter Web App, Twitter for Android, Twitter for iPhone, and several more. Table 1 explains the condition for the assignment of four such "source labels" by Twitter.

**Table 1.** Four Tweet Source Labels and the conditions for assignment of these labels by Twitter

| Tweet Source Label | Condition for Assignment |
|---|---|
| Twitter Web App | The tweet was posted by visiting the official website of Twitter [125] |
| Twitter for Android | The tweet was posted using the Twitter app for Android operating systems which is available for free download on the Google Playstore [126] |
| Twitter for iPhone | The tweet was posted using the Twitter app for iPhone, which is available for free download on the App Store [127] |
| TweetDeck | The tweet was posted by using TweetDeck, a social media dashboard application for the management of Twitter accounts [128] |

The Twitter platform allows its users to tweet in any of the 34 languages supported by Twitter [129]. Each of these languages is assigned a unique 2-letter code in the Twitter developer API, which helps to uniquely identify the associated language. Some examples include "en" for English, "ar" for Arabic, "bn" for Bengali, "cz" for Czech, "da" for Danish, "de" for German, "el" for Greek, "es" for Spanish, "fa" for Persian, and so on.

Inferring the type of tweet involved tracking the publicly available information about a tweet on Twitter that mentions whether it is an original tweet or a retweet (an original tweet that has been re-posted) or a reply (response to an original tweet or a retweet). Finally, detecting the domain embedded in a tweet involved processing the text associated with a tweet to detect any domains that may have been included in the tweet.

After detecting these characteristics and features from the set of 12028 tweets, in the next step, they were grouped together into distinct classes for quantification, categorization, and ranking. This helped to deduce multiple insights from each category of the results. For instance, in the tweet type category, this analysis helped to deduce the percentage of original tweets, retweets, and replies, which further helped in ranking these specific classes of results in the tweet type category. The ranking process helped to determine which of these respective classes constituted the maximum occurrence in each category. The above-mentioned functionalities for data mining, sentiment analysis, source tracking, language interpretation, type detection, and embedded URL analysis of Tweets were performed in a collective manner as per the flowchart shown in Figure 1 by using Social Bearing, as mentioned earlier in this section. The results that were obtained are discussed in Section 4.

### *3.4. Data Availability*

As described in Section 3.3, this work uses the Social Bearing research tool that works by complying with the Twitter standard search protocols. The policies of Twitter standard search state that – *"Keep in mind that the search index has a 7-day limit. In other words, no tweets will be found for a date older than one week"* [130]. Therefore, all the tweets that were analyzed in this study were posted in the range of May 5, 2022, and May 12, 2022 (the most recent date at the time of submission of this paper). However, for replication of this study, for repetition of this study on a bigger scale (by including tweets that were posted before May 5, 2022), and for the investigation of other research questions, we have developed an open-access dataset of Tweets posted about the Omicron variant since the first detected case of this variant on November 24, 2022.

This data collection was performed by the newly added Advanced Search feature provided by the Twitter platform [131] that allows searching of Tweets between any two given dates (the dates can be more than a week(s) or a month(s) apart) based on keyword search. As this is a feature developed by Twitter to support research and development using Twitter [132], it can be safely concluded usage of Twitter Advanced Search follows Twitter privacy policy and Twitter developer guidelines [103,104]. Furthermore, it can also be concluded that the process by which Twitter Advanced Search fetches tweets doesn't constitute any illegal, unethical, or unauthorized calling, posting, or scraping performed on the Twitter API [122].

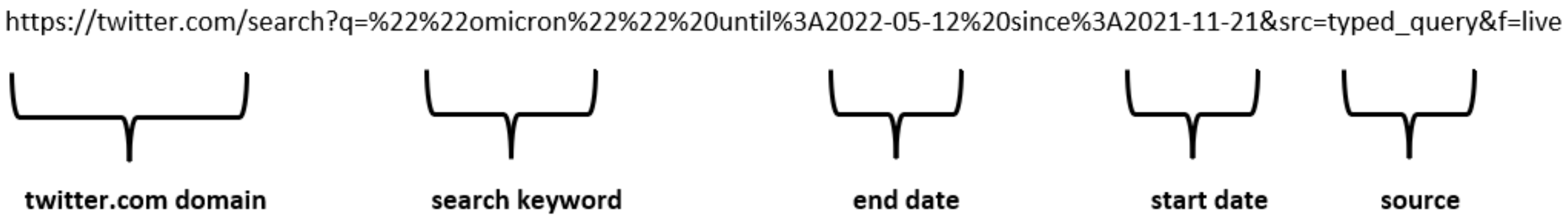

**Figure 2.** Representation of different parts of the RegEx that was used in Twitter Advanced Search.

The usage of Twitter Advanced Search involves logging into Twitter and then entering the relevant keywords and date range to develop a RegEx or regular expression. This RegEx is then used by the Twitter Advanced Search to display the relevant tweets as per the rules mentioned in the regular expression. This RegEx, along with the different parts of the same, that were used for this study is shown in Figure 2.

As can be seen from Figure 2, this RegEx comprised five parts. From left to right, in Figure 2, these parts represent the twitter.com domain, the search keyword, the end date, the start date, and the source. The search keyword was "omicron,", the end date was selected to be May 12, 2022, as this is the most recent date at the time of submission of this paper, and the start date was selected to be November 21, 2022, as, on this date, the first case of the Omicron variant was detected. The source parameter set as "typed_query" helped in the elimination of bot-related content where the underlining bots shared anything on Twitter that was not typed (such as images, advertisements, spam links, etc.). The results from the Twitter Advanced Search API were collected, compiled, and exported to develop this dataset. The dataset developed contains only Tweet IDs in compliance with the privacy policy, developer agreement, and guidelines for content redistribution of Twitter [103,104]. This section is divided into four subsections. The compliance of this dataset with the privacy policy, developer agreement, and guidelines for content redistribution of Twitter is described in sub-section 3.4.1. The compliance of this dataset with the FAIR (Findability, Accessibility, Interoperability, and Reusability) principles for scientific data management [132] is outlined in sub-section 3.4.2. Subsection 3.4.3 presents the data description and provides the URL that can be used to access and download the dataset. The instruction for using this dataset is presented in subsection 3.4.4. It is worth mentioning here that Twitter Advanced Search does not return an exhaustive list of Tweets posted between two specific dates. So, it is possible that multiple posted tweets posted between November 21, 2021, to May 12, 2022, were not returned by Twitter Advanced Search when the data collection was performed. Therefore, these tweets are not a part of this dataset. In addition to this, Twitter allows users the option to delete a tweet which would mean that there would be no retrievable Tweet text and other related information (upon hydration – which is explained in Section 3.4.4) for a Tweet ID of a deleted tweet. All the Tweet IDs available in this dataset correspond to tweets that have not been deleted at the time of writing this paper.

### *3.4.1. Compliance with guidelines for Twitter content redistribution*

The privacy policy of Twitter [103] states – *"Twitter is public and Tweets are immediately viewable and searchable by anyone around the world"*. The guidelines for Twitter content redistribution [104] state – *"If you provide Twitter Content to third parties, including downloadable datasets or via an API, you may only distribute Tweet IDs, Direct Message IDs, and/or User IDs (except as described below)*. It also states - *"We also grant special permissions to academic researchers sharing Tweet IDs and User IDs for non-commercial research purposes. Academic researchers are permitted to distribute an unlimited number of Tweet IDs and/or User IDs if they are doing so on behalf of an academic institution and for the sole purpose of non-commercial research."* Therefore, it may be concluded that mining relevant tweets from Twitter to develop a dataset (comprising only Tweet ID's) to share the same is in compliance with the privacy policy, developer agreement, and content redistribution guidelines of Twitter.

### *3.4.2 Compliance with FAIR*

The FAIR principles for scientific data management [132] state that a dataset should have Findability, Accessibility, Interoperability, and Reusability. The dataset is findable as it has a unique and permanent DOI. The dataset is accessible online. It is interoperable due to the use of .txt files for data representation that can be downloaded, read, and analyzed across different computer systems and applications. The dataset is re-usable as the associated tweets and related information such as user I.D., user name, retweet count, etc., for all the Tweet I.D.s can be obtained by the process of hydration in compliance with Twitter policies (Section 4), for data analysis and interpretation.

### *3.4.3. Data Description*

This section presents the description of the dataset publicly available at https://doi.org/10.5281/zenodo.6804323. The dataset consists of a total of 537,702 tweet ID's of tweets about the Omicron variant of COVID-19 that were posted on Twitter from November 21, 2021, to May 12, 2022. The Tweet ID's are presented in 7 different .txt files based on the timelines of the associated tweets. Table 2 provides the details of these dataset files.

**Table 2.** Description of all the files present in this dataset.

| Filename | No. of Tweet IDs | Date Range of the Tweet IDs |
|---|---|---|
| TweetIDs_November.txt | 17271 | November 21, 2021 to November 30, 2021 |
| TweetIDs_December.txt | 101393 | December 1, 2021 to December 31, 2021 |
| TweetIDs_January.txt | 95055 | January 1, 2022 to January 31, 2022 |
| TweetIDs_February.txt | 91571 | February 1, 2022 to February 28, 2022 |
| TweetIDs_March.txt | 100787 | March 1, 2022 to March 31, 2022 |
| TweetIDs_April.txt | 94409 | April 1, 2022 to April 20, 2022 |
| TweetIDs_May.txt | 37216 | May 1, 2022 to May 12, 2022 |

To comply with the privacy policy, developer agreement, and guidelines for content redistribution of Twitter [103,104], only the Tweet ID's associated with these 537,702 tweets are presented in this dataset. To obtain the detailed information associated with each of these tweets, such as the tweet text, user name, user I.D., timestamp, retweet count, etc., these Tweet ID's need to be hydrated. There are several applications, such as the Hydrator app [133], Social Media Mining Toolkit [134], and Twarc [135], that work by complying with Twitter policies and may be used for hydrating the Tweet ID's in this dataset. A step-by-step process for using one of these applications – the Hydrator app for hydrating the files in this dataset, is presented in Section 3.4.4.

#### *3.4.4. Instructions for using the Dataset*

The following is the step-by-step process for using one of these applications, the Hydrator app [133], to hydrate this dataset or, in other words, to obtain the text of the tweet, user ID, user name, retweet count, language, tweet URL, source, and other public information related to all the Tweet IDs present in this dataset. The Hydrator app works in compliance with the policies for accessing and calling the Twitter API.

1. Download and install the desktop version of the Hydrator app from https://github.com/DocNow/hydrator/releases.
2. Click on the "Link Twitter Account" button on the Hydrator app to connect the app to an active Twitter account.
3. Click on the "Add" button to upload one of the dataset files (such as TweetIDs_November.txt). This process adds the dataset file to the Hydrator app.
4. If the file upload is successful, the Hydrator app will show the total number of Tweet I.D.'s present in the file. For instance, for the file - "Tweet-IDs_November.txt ", the app would show the Number of Tweet IDs as 17271.
5. Provide details for the respective fields: Title, Creator, Publisher, and URL in the app, and click on "Add Dataset" to add this dataset to the app.
6. The app would automatically redirect to the "Datasets" tab. Click on the "Start" button to start hydrating the Tweet IDs. During the hydration process, the progress indicator would increase, indicating the number of Tweet IDs that have been successfully hydrated and the number of Tweet IDs that are pending hydration.
7. After the hydration process ends, a .jsonl file would be generated by the app that the user can choose to save on the local storage.
8. The app would also display a "CSV" button in place of the "Start" button. Clicking on this "CSV" button would generate a .csv file with detailed information about the tweets, which would include the text of the tweet, user ID, user name, retweet count, language, tweet URL, source, and other public information related to the tweet.
9. Repeat steps 3-8 for hydrating all the files of this dataset.

## 4. Results and Discussions

This section presents and discusses the results obtained upon the development and implementation of the proposed methodology on the set of 12028 tweets about the SARS-CoV-2 omicron variant posted on Twitter from May 5, 2022, to May 12, 2022. The results are shown in Figures 3-7. Figure 3 shows the results of sentiment analysis. The specific categories into which the sentiment of a tweet was classified comprised 'great', 'good', 'neutral', 'bad', and 'terrible'. As can be seen from Figure 3, the 'neutral' emotion was present in a majority of the tweets (50.5% of the total tweets). It was followed by tweets that had the 'bad' (15.5% of the total tweets) and 'good' (14.0% of the total tweets) emotions associated with them. These respective sentiment categories were followed by the sentiment categories of 'terrible' (12.5% of the total tweets) and 'great' (7.5% of the total tweets).

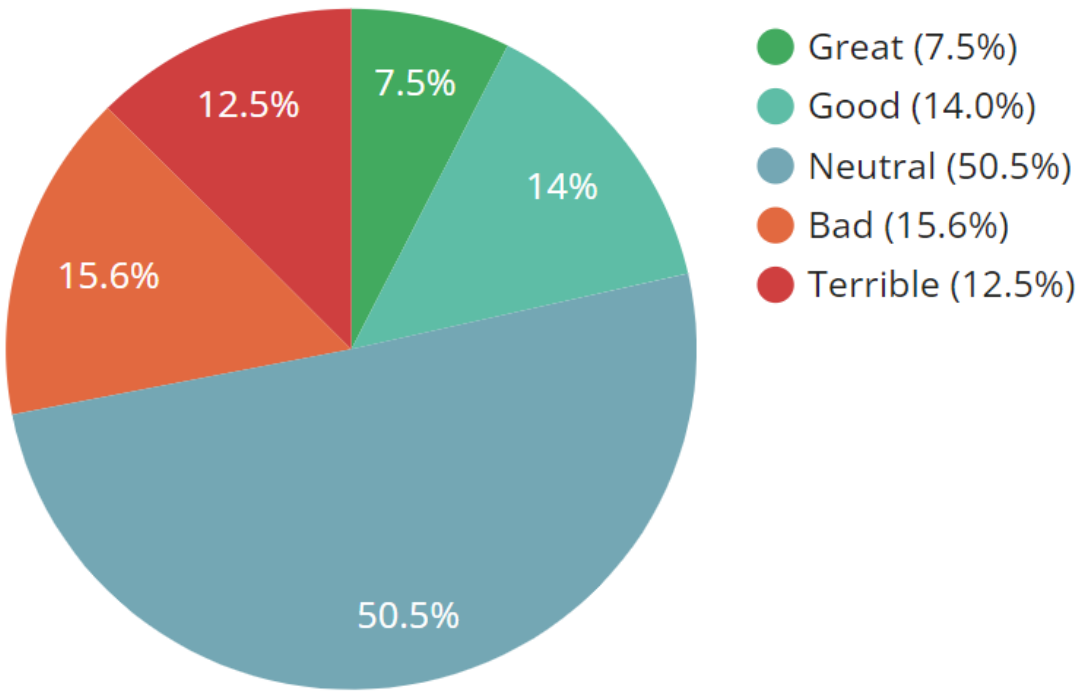

**Figure 3.** Results of sentiment analysis performed on tweets about the SARS-CoV-2 omicron variant.

The results of the Tweet Type detection are shown in Figure 4. This process involved detection, categorization, and ranking of the tweets into original tweets, retweets, and replies. In the index shown in this Figure, original tweets are referred to as "tweets". As can be seen from this Figure, retweets comprised a majority (60.8%) of all the tweets about the SARS-CoV-2 omicron variant. It was followed by original tweets (19.8% of the total tweets) and replies (19.4% of the total tweets).

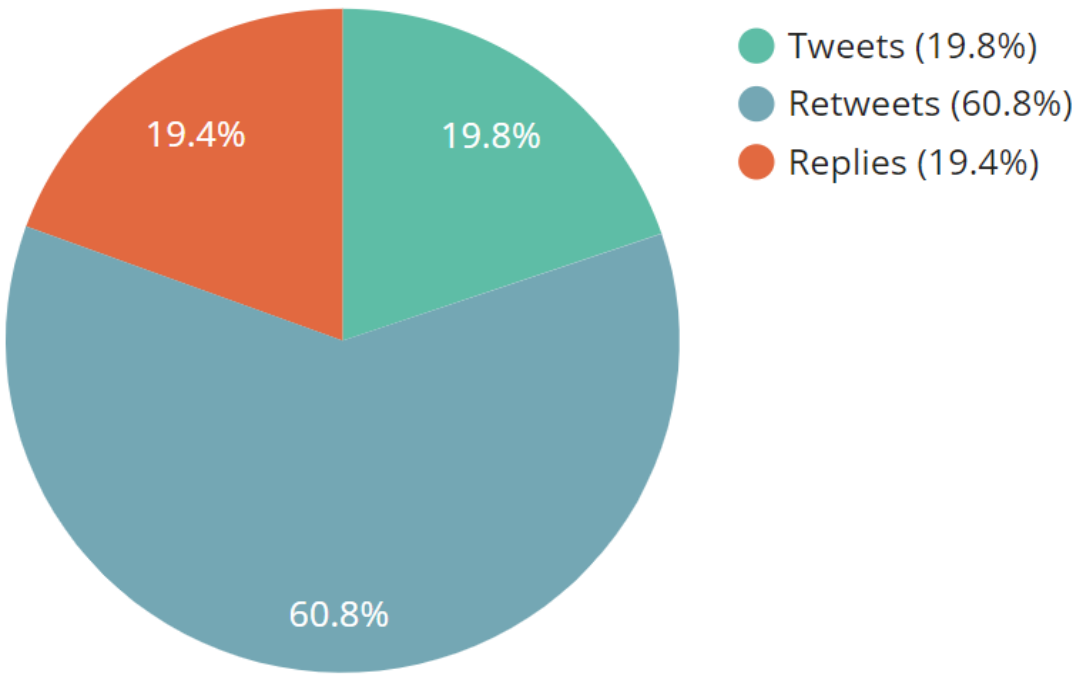

**Figure 4.** Results of Tweet type detection performed on tweets about the SARS-CoV-2 omicron variant.

Figure 5 shows the results of tweet source detection as publicly shown by the Twitter platform. Here, "Android" refers to "Twitter for Android", "iPhone" refers to "Twitter for iPhone", and "iPad" refers to "Twitter for iPad" (meanings of these categories are mentioned in Table 2). The total number of sources in this corpus of 12028 tweets was observed to be more than 10; the top 10 sources are listed in the index provided in this Figure for clarity and readability. As can be seen from this Figure, "Twitter for Android" accounted for the most number of tweets (35.2% of the total tweets). It was followed by "Twitter Web App" (29.2% of the total tweets), "Twitter for iPhone" (25.8% of the total tweets), "Twitter for iPad" (3.8% of the total tweets), and "TweetDeck" (1.6% of the total tweets). "TweetDeck" was followed by a few other sources, which accounted for less than 1% of the total tweets.

The results of the language detection and ranking of the used languages are shown in Figure 6. The total number of languages detected from all the 12028 tweets was observed to be more than 10; the top 10 languages are listed in the index provided in this Figure for clarity and readability. As can be seen from this Figure, more than a majority (65.9%) of the tweets were written in English. In the second place was Spanish or Castilian (10.5% of the total tweets), which was followed by French (5.1% of the total tweets), Italian (3.3% of the total tweets), Japanese (2.5% of the total tweets), and a few other sources that accounted for less than 2% of the total tweets.

The results of the detection of the embedded URLs and the ranking of the associated clusters are shown in Figure 7. The total number of different domains in this corpus of 12028 tweets was observed to be more than 10; the top 10 embedded domains are listed in the index provided in this Figure for clarity and readability. As can be seen from Figure 6, the domain – twitter.com comprised the highest count. This can be attributed to the fact that retweets comprised a significant percentage of the tweets about the SARS-CoV-2 omicron variant. It was followed by the domains - biorxiv.org, nature.com, wapo.st, nzherald.co.nz, recvprofits.com, science.org, bit.ly, YouTube (youtu.be is a shortened version of YouTube URLs [136]), and sciencedirect.com.

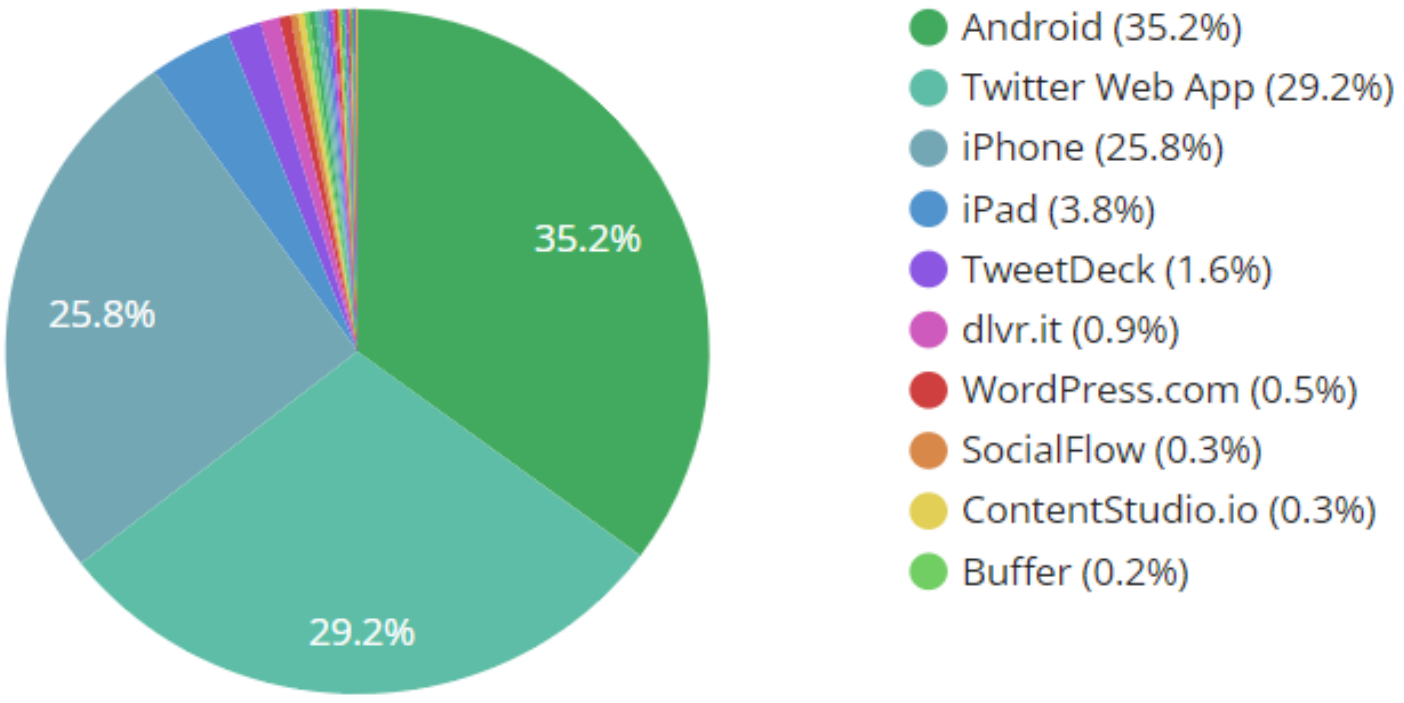

**Figure 5.** Results of Tweet source detection performed on tweets about the SARS-CoV-2 omicron variant

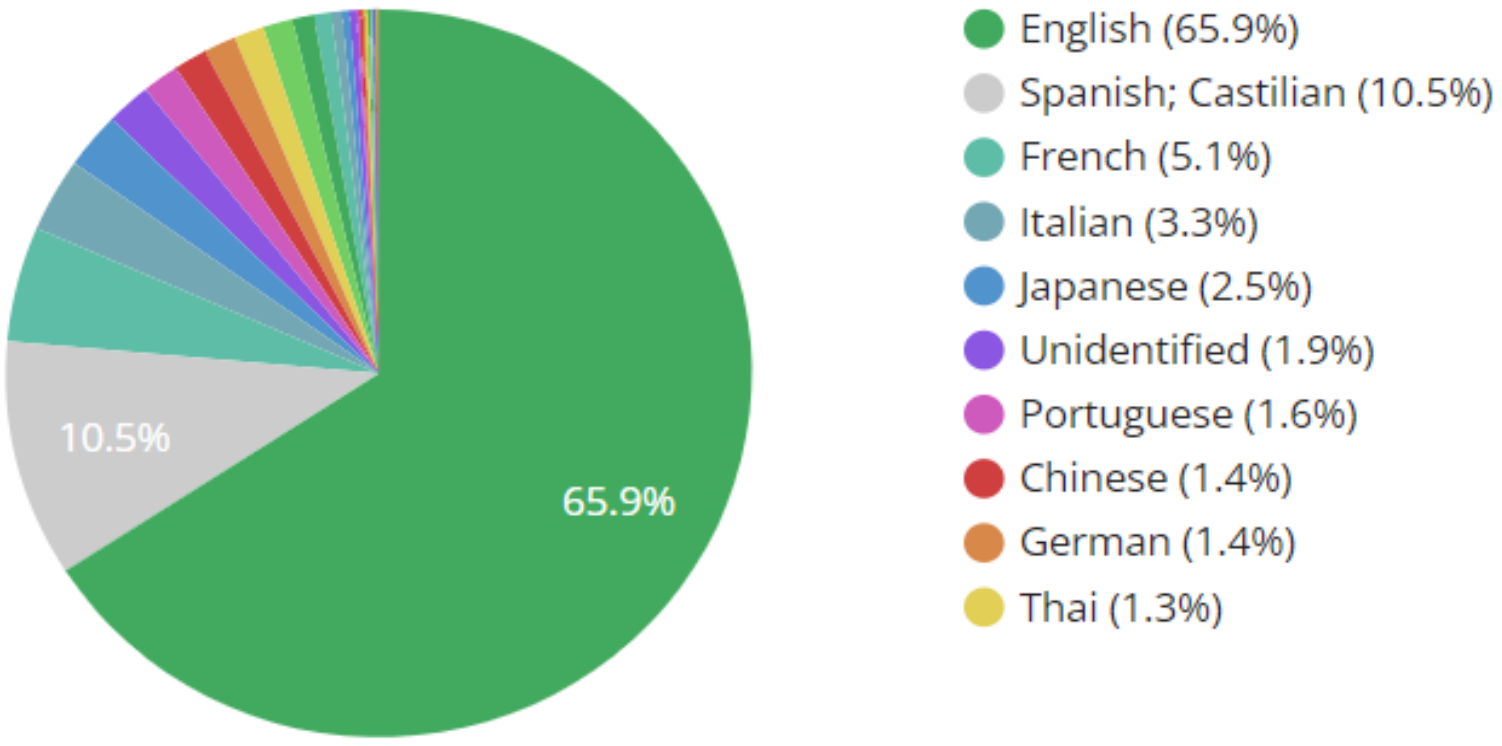

**Figure 6.** Results of language detection performed on tweets about the SARS-CoV-2 omicron variant

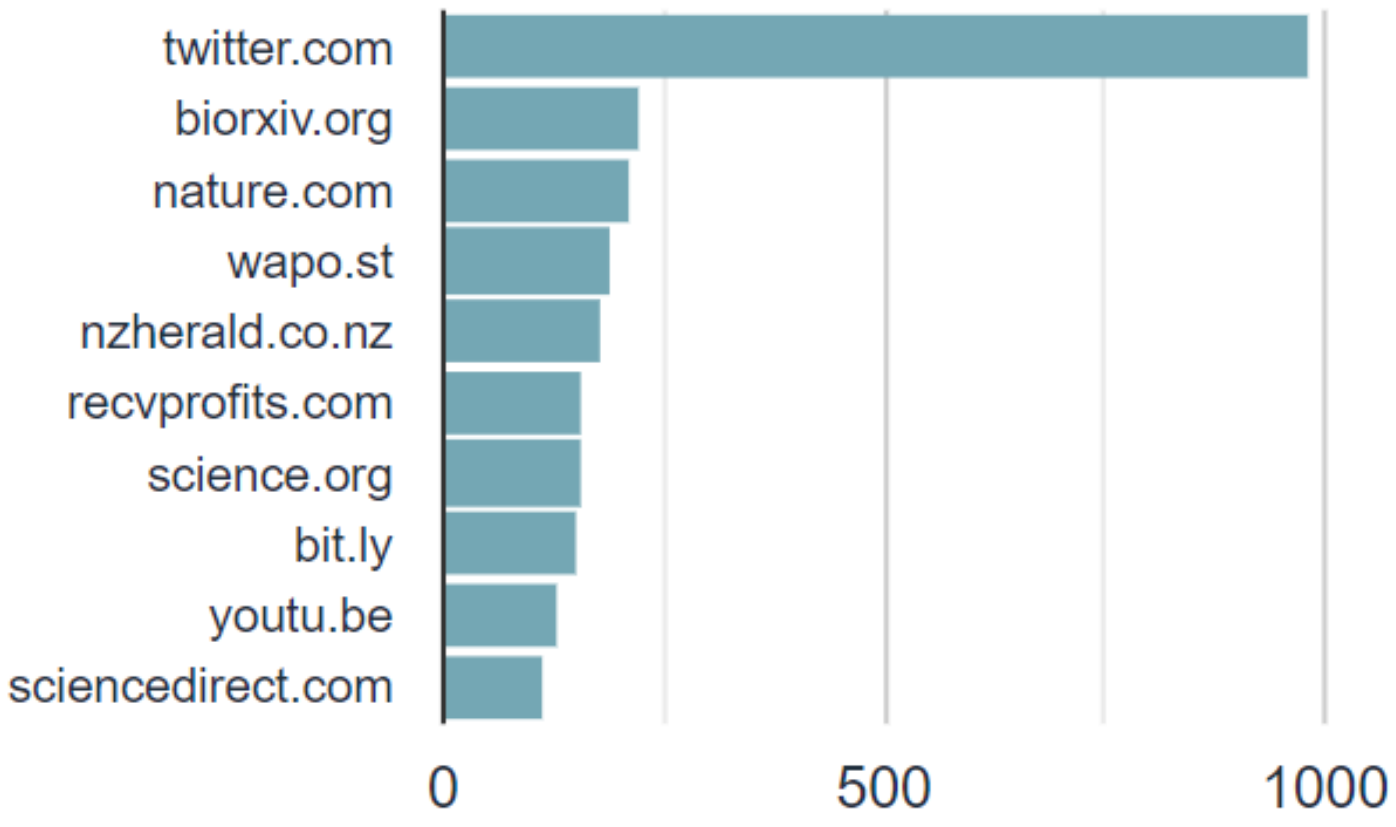

**Figure 7.** Results of embedded URL detection performed on tweets about the SARS-CoV-2 omicron variant

In addition to the above, a keyword-based word cloud analysis and a hastag-based word cloud analysis were performed on this set of 12028 tweets about the SARS-CoV-2 omicron variant. The results are shown in Figures 8 and 9, respectively. In each of these word clouds, a different font color has been used to identify a keyword and hashtag, respectively. The font size of each of these words and hashtags is directly proportional to the frequency of the same. Or in other words, the keyword (or hashtag) that has the highest frequency has the largest font size, and the keyword (or hashtag) that has the lowest frequency has the smallest font size. Those keywords (or hashtags) which had a very low frequency and would subsequently have an extremely small font size were not included in the results shown to avoid cluttering and to enhance the readability of these Figures. As can be seen from both these figures, the "omicron" keyword was the most frequent keyword, and similarly, the "#omicron" hashtag was the most frequent hashtag across all these tweets. This observation further helps to support the proposed approach for searching tweets by using "omicron" as the keyword or hashtag to be searched, which was outlined at the beginning of this section.

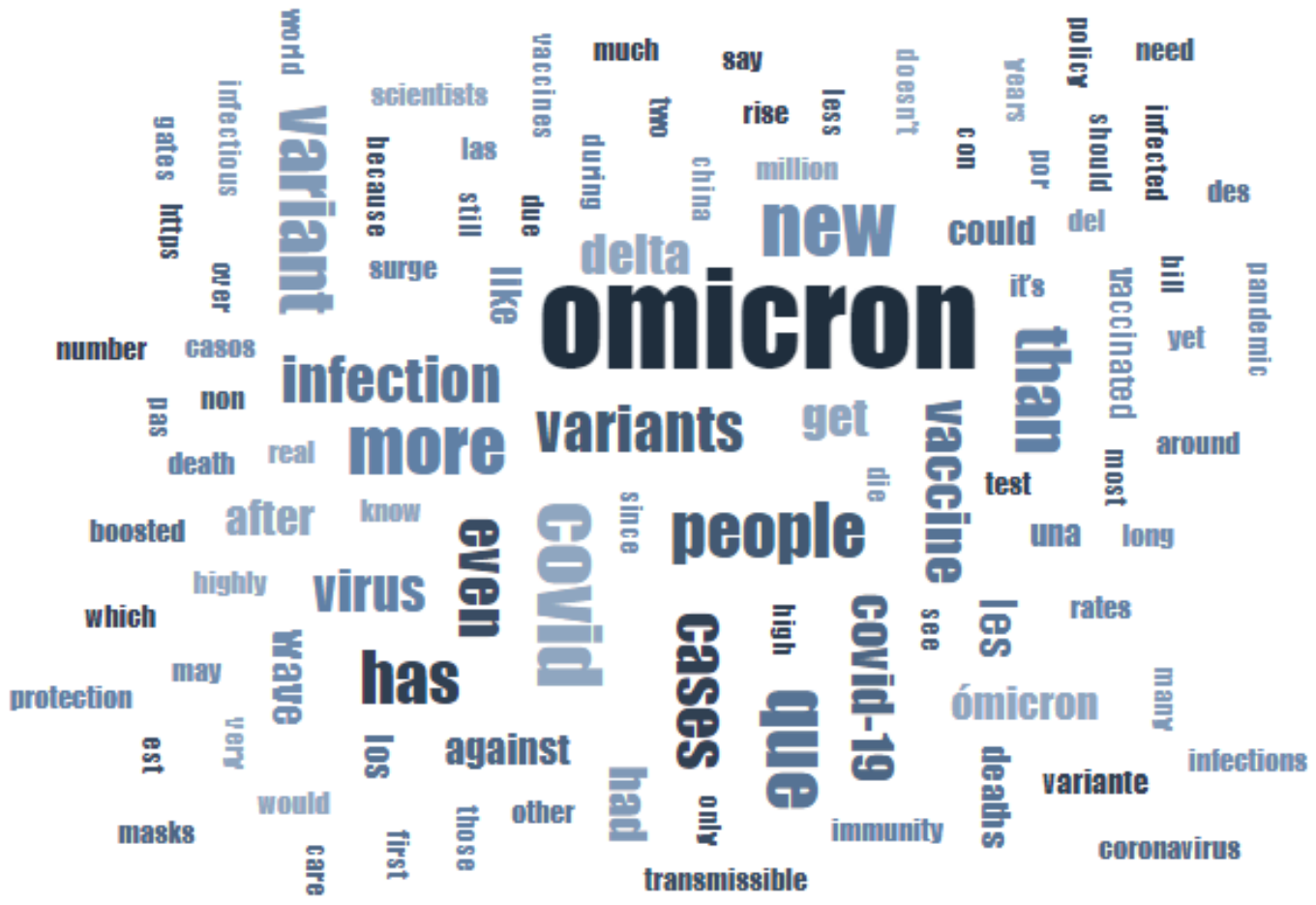

**Figure 8.** Results of word cloud analysis of the keywords present in the tweets about the SARS-CoV-2 omicron variant

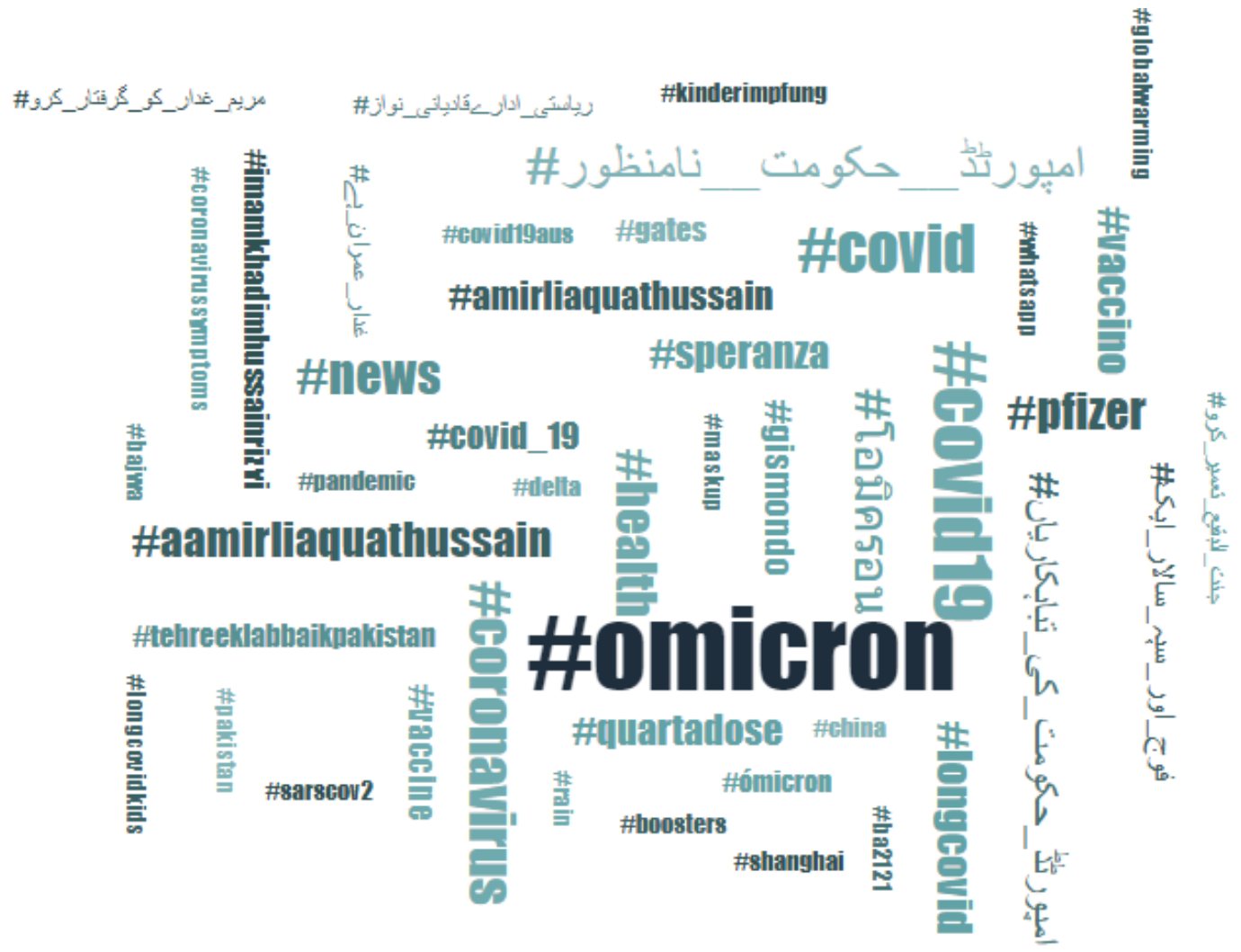

**Figure 9.** Results of word cloud analysis of the hashtags present in the tweets about the SARS-CoV-2 omicron variant

In addition to these findings, a discussion is presented next that upholds how this work addresses the multiple distinct needs in this field of research that were identified after a comprehensive review of recent works (presented in Section 2).

1. The previous works in this field proposed approaches to filter tweets based on one or more keywords, hashtags, or phrases such as "COVID-19", "coronavirus", "SARS-CoV-2", "covid", "corona" but did not contain any keyword or phrase specifically related to the omicron variant. So, those approaches for tweet searching or tweet filtering might not be applicable to the tweets posted about the omicron variant unless the Twitter user specifically mentions something like "COVID-19 omicron variant" or "SARS-CoV-2 Omicron variant" in their tweets. As discussed in Section 3, there were multiple instances when the Twitter users did not use keywords, hashtags, or phrases such as "COVID-19", "coronavirus", "SARS-CoV-2", "covid", "corona" along with the keyword or hashtag – "omicron". Thus, the need is to develop an approach to specifically mine tweets posted about the omicron variant. This work addresses this need by proposing a methodology that searches tweets based on the presence of "omicron" either as a keyword or as a hashtag. The effectiveness of this approach is justified by the word clouds presented in Figures 8 and 9.

2. The prior works [88,89] on sentiment analysis in the context of tweets about COVID-19 focused on developing approaches for classifying the sentiment only into three classes - 'positive', 'negative', and 'neutral'. In a realistic scenario, there can be different kinds of 'positive' emotions, such as 'good' and 'great'. Similarly, there can be different kinds of 'negative' emotions, such as 'bad' and 'terrible'. The existing works cannot differentiate between these kinds of positive (or negative) emotions. To address the need in this context, associated with increasing the number of classes for classification of the sentiment of the tweet, this work proposes an approach that classifies tweets into five sentiment classes: 'great', 'good', 'neutral', 'bad' and 'terrible.'
3. The emerging works in this field, for instance, related to detecting trending topics [90], anomaly events [91], public perceptions towards C.D.C. [92], and views towards not wearing masks [93], focused on the development of new frameworks and methodologies without focusing on quantifying multimodal components of the characteristics of the tweets and ranking these characteristics to infer insights about social media activity on Twitter due to COVID-19. This work addresses this need. The results from sentiment analysis, type detection, source tracking, language interpretation, and embedded URL observation were categorized into distinct categories, and these categories were ranked in terms of the associated characteristics to infer meaningful and relevant insights about social media activity on Twitter related to tweets posted about the SARS-CoV-2 Omicron variant (Figures 3-7). For instance, in the tweet type analysis, the findings of this study show that "Twitter for Android" accounted for the most number of tweets (35.2% of the total tweets), which was followed by "Twitter Web App" (29.2% of the total tweets), "Twitter for iPhone" (25.8% of the total tweets), and other sources.
4. The previous works centered around performing data analysis on tweets related to COVID-19 included a small corpus of tweets, such as 4081 tweets in [102] and 4492 tweets in [100]. In view of the number of active Twitter users and the number of tweets posted each day, there is a need to include more tweets in the data analysis process. This work addresses this need by considering a total of 12028 relevant tweets that had a combined reach of 149,500,959, 226,603,833 impressions, 1,053,869 retweets, and 3,427,976 favorites.
5. The development of Twitter datasets has been of significant importance and interest to the scientific community in the areas of Big Data Mining, Data Analysis, and Data Science. This is evident from the recent Twitter datasets on 2020 U.S. Presidential Elections [72], 2022 Russia Ukraine war [73], climate change [74], natural hazards [75], European Migration Crisis [76], movies [77], toxic behavior amongst adolescents [78], music [79], civil unrest [80], drug safety [81], and Inflammatory Bowel Disease [82]. Twitter datasets help to serve as a data resource for a wide range of applications and use cases. For instance, the Twitter dataset on music [79] has helped in the development of a context-aware music recommendation system [137], next-track music recommendations as per user personalization [138], session-based music recommendation algorithms [139], music recommendation systems based on the use of affective hashtags [140], music chart predictions [141], user-curated playlists [142], sentiment analysis of music [143], listener engagement with popular songs [144], culture aware music recommendation systems [145], mining of user personalities [146], and several other applications. The works related to the development of Twitter datasets on COVID-19 [83-87] in the last few months did not focus on the development of a Twitter dataset comprising Tweets about the Omicron variant of COVID-19 since the first detected case of this variant. To address this research gap, we developed a Twitter dataset comprising 537,702 Tweet IDs of the same number of Tweets about the Omicron variant of COVID-19 since the first detected case of this variant on November 24, 2021. As per the best knowledge of the authors, such a comprehensive analysis of Tweets about the Omicron variant of COVID-19 has not been done before. The results presented and discussed support the novelty and relevance of this research work. However, the paper has a few limitations. First, the research tool – Social Bearing that was used uses several JavaScript libraries which include - jquery-ui.min.js [106], jquery.tinysort.min.js [107], masonry.pkgd.min.js [108], d3.layout.cloud.js [109], d3.min.js [110], analytics.js [111], and loader.js [112] for performing multimodal analysis of Tweets as presented in this paper. However, we did not develop any algorithm or software solution of our own to perform the same analysis and/or verify the results presented by this research tool. If any algorithm or software solution is developed that uses the same methodology of integrating the functionality of these JavaScript libraries as the Social Bearing research tool, then it is likely that the results would be identical. However, use of other methodologies such as Bidirectional Long Short-Term Memory (Bi-LSTM), Bidirectional Encoder Representations from Transformers (BERT), Word Vector Analysis, Topic Modeling, Lexicon-based analysis, Dialogue Bidirectional Encoder Representations from Transformers (DialBERT), and Deep Learning, in such an algorithm or software solution might provide slightly different results in terms of the number of entities classified into each bin during the classification process. For instance, our approach shows that 50.5% of the tweets had a 'neutral' emotion. Any of these alternate approaches might provide a number very close to 50.5% but not exactly equal to 50.5%. To address this limitation, we plan to develop, implement, and test all these alternative approaches on this dataset to identify the best approach for performing such a comprehensive analysis on Tweets about the SARS-CoV-2 omicron variant. Second, as the SARS-CoV-2 omicron variant continues to spread across the globe, rapid advances are being made related to omicron-specific vaccines [147,148], and new studies related to this variant are also getting published [149, 150]. These advances, studies, new findings, and the nature of public reaction, views, opinion, feedback, thoughts, and perspectives towards the same may impact one or more of the characteristic features of social media behavior on Twitter that were investigated and analyzed in this study. To address this limitation, a follow-up study will be conducted in the future. In that follow-up study, the aim would be to include all the relevant tweets in between the dates when the first case of Omicron was recorded and when the last case of Omicron would be recorded to perform this exploratory analysis once again to compare the associated findings and to comment on any similarities and dissimilarities in the insights that might be observed. Third, the in-built feature of Social Bearing that was used to remove bot-related content might not be the most optimal approach for removal of all bot-related tweets as certain advanced bot accounts that exactly mimic real user Twitter accounts in terms of Twitter user names and share/post content on Twitter after randomized time intervals (as opposed to the usual bots which usually share/post content after a fixed time interval) might be difficult to identify using this approach. Emerging works in the field of Natural Language Processing such as TweezBot [151], Bot-DenseNet [152], Bot2Vec [153], Botter [154], and GlowWorm-based Generalized Regression [155] could be used for identifying such advanced bot accounts on Twitter. We could not implement any of these emerging works in our study due to the limited integration options provided by the Social Bearing research tool. We plan to address this limitation in the future by developing our own algorithm or software solution that would not have any such limited integration options.

## 5. Conclusions

Since the initial outbreak in Wuhan, China, in December 2019, the SARS-CoV-2 virus has resulted in a total of 520,628,068 cases and 6,287,207 deaths on a global scale. The virus has undergone multiple mutations, and as a result, several variants have been detected, such as Alpha, Beta, Gamma, Delta, Epsilon, Eta, Iota, Kappa, Zeta, Mu, and Omicron, in different parts of the world. Out of these variants, the Omicron variant, a variant of concern (VOC) as per WHO, is currently the globally dominant variant and has been reported to be the most immune evasive VOC of SARS-CoV-2 and is also considered to be highly resistant against antibody-mediated neutralization. The number of cases and deaths due to Omicron in different parts of the world is on a constant rise.

Research conducted during pandemics in the past suggests that people extensively use social media platforms for communication, information, news, views, opinions, ideas, knowledge, feedback, and experiences related to the pandemic they are facing. Twitter, one such social media platform, is popular amongst all age groups. Therefore, this work took a comprehensive approach to identify, study, and analyze tweets related to the SARS-CoV-2 Omicron variant to understand, categorize, and interpret the associated dynamics and characteristic features of social media behavior. A total of 12028 tweets about the SARS-CoV-2 Omicron variant were mined from Twitter, and the associated sentiment ('great', 'good', 'neutral', 'bad', and 'terrible'), type (original tweet, retweet, or reply), source (such as "Twitter for Android", "Twitter Web App", "Twitter for iPhone", etc.), language (such as English, Spanish, Bengali, etc.), and embedded URLs (URLs that were included in the tweet text) were analyzed. The findings from this exploratory study are manifold. First, a majority of the tweets had a 'neutral' emotion (50.5% of the total tweets), which was followed by 'bad' (15.6% of the total tweets), 'good' (14.0% of the total tweets), 'terrible' (12.5% of the total tweets), and 'great' (7.5% of the total tweets). Second, 35.2% of the tweets had "Twitter for Android" as their source. It was followed by the "Twitter Web App" (29.2% of the total tweets), "Twitter for iPhone" (25.8% of the total tweets), "Twitter for iPad" (3.8% of the total tweets), "Tweetdeck" (1.6% of the total tweets), and other sources which accounted for less than 1% of the total tweets. Third, a majority of the tweets (65.9%) were posted in English, which was followed by Spanish or Castillian (10.5% of the total tweets), French (5.1% of the total tweets), Italian (3.3% of the total tweets), Japanese (2.5% of the total tweets), and other languages that accounted for less than 2% of the tweets. Fourth, the majority of the tweets (60.8%) in terms of tweet type were retweets which were followed by original tweets (19.8% of the total tweets) and replies (19.4% of the total tweets). Fifth, in terms of embedded domains, the most common domain embedded in the tweets was found to be twitter.com, which was followed by biorxiv.org, nature.com, wapo.st, nzherald.co.nz, recvprofits.com, science.org, and a few other sources. Finally, to support research and development in this field, this work presents an open-access Twitter dataset of 537,702 Tweet IDs of the same number of Tweets about the SARS-CoV-2 omicron variant that was posted on Twitter since the first detected case of this variant on November 24, 2022. Future work would involve addressing the limitations of this study as discussed at the end of the previous section. In addition to this, the dataset associated with this work would be updated on a routine basis so that the research community has access to the most recent data in this regard.

**Author Contributions:** Conceptualization, NT; methodology, N.T.; software, N.T.; validation, N.T.; formal analysis, N.T.; investigation, N.T.; resources, N.T.; data curation, N.T.; visualization, N.T.; data analysis and results, N.T.; writing—original draft preparation, N.T.; writing—review and editing, N.T.; supervision, Not Applicable; project administration, C.Y.H.; funding acquisition, Not Applicable. All authors have read and agreed to the published version of the manuscript.

**Supplementary Materials:** Not applicable.

**Funding:** This research received no external funding

**Institutional Review Board Statement:** Not applicable

**Informed Consent Statement:** Not applicable.

**Data Availability Statement:** The data presented in this study are publicly available at https://doi.org/10.5281/zenodo.6804323

**Conflicts of Interest:** The author declares no conflict of interest